\documentclass[final,3p,times,twocolumn,numbers]{elsarticle}

\usepackage{subfigure}

\usepackage{amssymb}
\usepackage{amsmath}
\journal{Pattern Recognition Letters}

\begin{document}

\begin{frontmatter}



\title{Segmenting the Complex and Irregular in Two-Phase Flows: \\A Real-World Empirical Study with SAM2}

\author[label1]{Semanur Küçük}
\ead{S.Kuecuek@tudelft.nl}

\author[label1]{Cosimo Della Santina}
\ead{C.DellaSantina@tudelft.nl}

\author[label1]{Angeliki Laskari}
\ead{A.Laskari@tudelft.nl}

\affiliation[label1]{organization={Delft University of Technology, Faculty of Mechanical Engineering},
             addressline={Mekelweg 5},
             city={Delft},
             postcode={2628 CD},
             country={Netherland}}

\begin{abstract}
Segmenting gas bubbles in multiphase flows is a critical yet unsolved challenge in numerous industrial settings, from metallurgical processing to maritime drag reduction. Traditional approaches—and most recent learning-based methods—assume near-spherical shapes, limiting their effectiveness in regimes where bubbles undergo deformation, coalescence, or breakup. This complexity is particularly evident in air lubrication systems, where coalesced bubbles form amorphous and topologically diverse patches. In this work, we revisit the problem through the lens of modern vision foundation models. We cast the task as a transfer learning problem and demonstrate, for the first time, that a fine-tuned Segment Anything Model (SAM v2.1) can accurately segment highly non-convex, irregular bubble structures using as few as 100 annotated images. 
\end{abstract}




\end{frontmatter}

\section{Introduction}

Accurate segmentation of bubbles or air patches from optical measurements plays a crucial role in analyzing two-phase flows, as it underpins the study of drag reduction, turbulence modulation, and interfacial dynamics \cite{tanaka2023downstream, ni2024deformation,wang2023experimental}. However, this important task remains challenging due to several factors, including overlapping bubble boundaries, inconsistent lighting conditions, image noise, and irregular bubble shapes that deviate from ideal spherical forms. Qin et al. \cite{qin2017stream} demonstrate that standard segmentation algorithms \cite{serra2020two, ronneberger2015u, schmidt2018cell, he2017mask} often fail to provide accurate results, forcing researchers to manually inspect and refine the segmentation, thus substantially limiting the scalability of the analyses. 

To overcome these limitations and motivated by recent advances in computer vision, bubble detection research has increasingly shifted toward deep learning, in the hope that these techniques can better handle complex scenarios.
Early efforts by Ilonen et al. and Serra et al. explored the application of flat ANN to bubble segmentation, establishing baselines for later data-driven methods \cite{ilonen2018comparison, serra2020two}. This was followed by the adoption of vanilla convolutional neural networks (CNNs), with Soibam et al. and Malakhov et al. targeting boiling flows under constrained conditions \cite{soibam2023application, malakhov2023deep}, and Kim and Park extending the analysis to varying flow regimes through a dedicated network design \cite{kim2021deep}. To extend segmentation performance beyond tightly controlled conditions, researchers have explored more advanced network architectures. Hessenkemper et al. \cite{hessenkemper2022bubble} compared U-Net \cite{ronneberger2015u}, StarDist \cite{schmidt2018cell}, and Mask R-CNN \cite{he2017mask}, finding that a hybrid of U-Net and StarDist yielded the most robust results across variable scenarios. In parallel, Haas et al. \cite{haas2020bubcnn} introduced BubCNN, a composite model combining Faster R-CNN \cite{ren2015faster} with a shape regression module trained on over 100,000 annotated bubbles. Still, even with extensive training data, these models struggled with dense bubble clusters, elevated void fractions, and non-uniform lighting conditions, which are commonly encountered in bubbly datasets.
\begin{figure*}[th!]    \centering    \includegraphics[width=\textwidth]{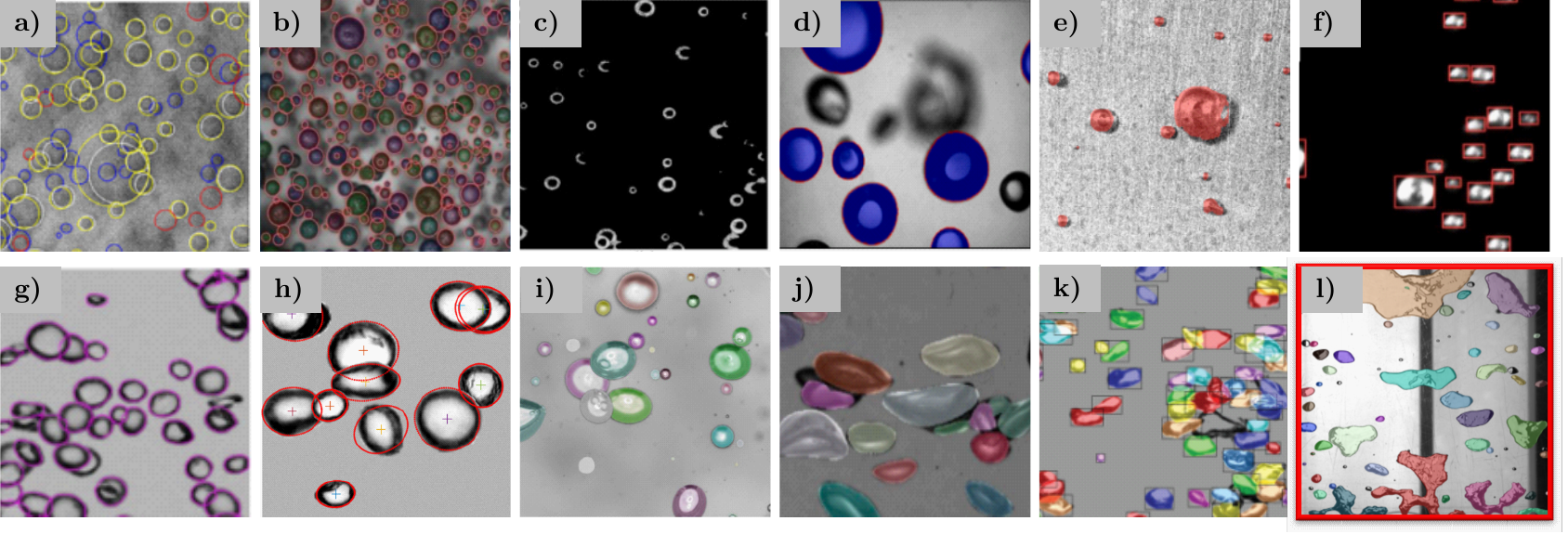}    \caption{Segmentation results from various deep learning-based approaches used in two-phase flow studies, ordered from round to increasingly deformable morphologies. (a) classical vs data-driven comparison by Ilonen et al. (2018) \cite{ilonen2018comparison}, (b) transfer learning with COCO-pretrained Mask R-CNN by Cui et al. (2022) \cite{cui2022deep},(c) hybrid RHT + Neural Network method by Serra et al. (2020) \cite{serra2020two}, (d) Malakhov et al. (2023) under varying pressures \cite{malakhov2023deep},(e) CNN-based segmentation by Soibam et al. (2023) \cite{soibam2023application}, (f) real-time segmentation with SAM-assisted YOLO by Wang et al. (2025) \cite{wang2025quantifying}, (g) universal CNN model by Kim and Park (2021) \cite{kim2021deep}, (h) BubCNN model using Faster R-CNN and shape regression by Haas et al. (2020) \cite{haas2020bubcnn}, (i) BubSAM model based on the Segment Anything Model (SAM) by Xu et al. (2024) \cite{xu2024bubsam}, (j) comparative study of U-Net, StarDist, and Mask R-CNN by Hessenkemper et al. (2022) \cite{hessenkemper2022bubble}, (k) low-data transfer learning using ResNet-50 and synthetic images by Homan and Deen (2024) \cite{homan2024deep}, and (l) segmentation output from the model proposed in this study.}    \label{fig:hist}\end{figure*}

In recent years, transfer learning has emerged as a potential solution to this additional challenge.
Cui et al. fine-tuned a COCO-pretrained Mask R-CNN on just 70 images, achieving accurate results up to 14.7\% gas holdup \cite{cui2022deep}. Homan and Deen used synthetic single-bubble masks to adapt a ResNet-50-based Mask R-CNN for use on modest hardware \cite{homan2024deep}. Wang et al. explored SAM-assisted pipelines for real-time segmentation \cite{wang2025quantifying}, and Xu et al. provided the first systematic evaluation of SAM’s capabilities for this task \cite{xu2024bubsam}, while Ali et al. \cite{khojasteh2024practical} also reported promising SAM-based results on bubbly flows. Among these, SAM-based approaches stand out as particularly promising—despite current challenges with overlapping mask generation.

Still, despite this substantial recent progress, a key limitation persists, as illustrated in Figure~\ref{fig:hist}: most studies focus on nearly spherical, isolated bubbles, often confined to narrow size ranges. This reduces variability and simplifies the learning task, but fails to capture the complex, deformable morphologies that characterize real-world multiphase flows.

In this work, we present an empirical investigation of bubble segmentation under complex, real-world conditions. We evaluate SAM 2.1 for the first time on a bubble segmentation task, benchmarking its ability to segment dense, irregular, and size-varying bubble structures. Our dataset spans a size distribution several orders of magnitude wider than those used in prior work and includes bubbles ranging from perfectly spherical to highly non-convex and topologically complex shapes (Fig. \ref{fig:hist} l)). We further investigate how fine-tuning and data augmentation impact performance in these challenging settings. As a byproduct, we publicly release the labeled dataset used in our study, aiming to support future research in advancing bubble segmentation beyond the simplified regime of isolated, near-spherical bubbles.

\section{Methodology}
\subsection{Dataset}

The dataset used in this study originates from prior experimental work by one of the authors \cite{Laskari2025}, focused on air lubrication flows. This case was chosen for its practical relevance and the inherent complexity of the air phase topology, including not only discrete bubbles but also merged, elongated, and irregular air patches that deform continuously under turbulent flow conditions.
The experiments were performed in a turbulent boundary layer flow over a flat plate fitted with a slot-type air injector. A Phantom 640-L high-speed camera, equipped with a 105 mm lens, was positioned above the plate to capture the air phase at 500 Hz. Illumination was provided by two LED panels ensuring high-contrast images suitable for detecting bubbles. The complete dataset contains several thousand images, recorded under varying flow conditions. 

For this work, we selected 350 images corresponding to the bubbly flow regime and manually annotated each of them. To ensure temporal independence, we down-sampled time resolved sets to 5 Hz, minimizing the likelihood that the same bubble appears in multiple frames.
%
%
We generate masks that closely follow the outlines of the bubbles, rather than relying on bounding boxes. Note indeed that capturing detailed shape and contour information is essential for tasks such as deformation analysis and centroid estimation. Figure ~\ref{fig:result2} a) illustrates a representative annotation from the dataset, for which the average gas hold-up is 19.5\%.



\subsection{Bubble Categorization}

\begin{figure}[t]
    \centering
    \includegraphics[width=\columnwidth]{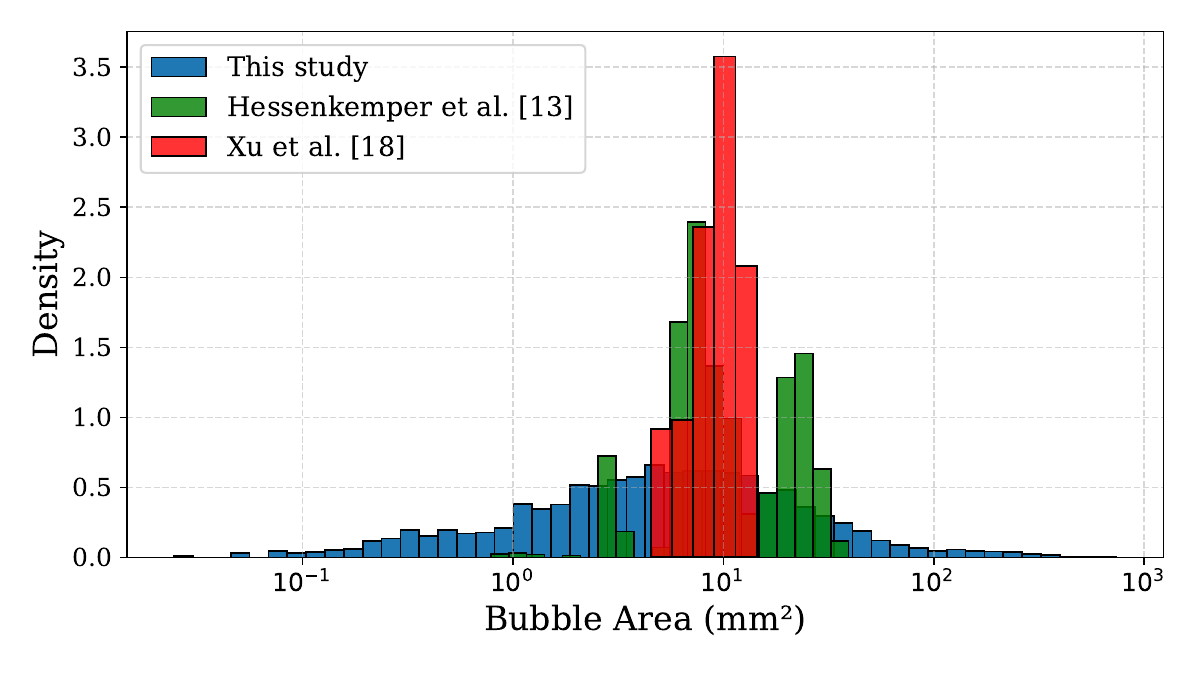}\vspace{-0.35cm}
    \caption{{Probability density function of the (wall-projected) area of all identified bubbles, including the current dataset (blue) compared with data from \cite{hessenkemper2022bubble} and \cite{xu2024bubsam}; broader distributions indicate increased multi-dispersity in air bubble size.}}
    \label{fig:size_comparison}
\end{figure}

\begin{figure}[t]
    \centering
    \includegraphics[width=\columnwidth]{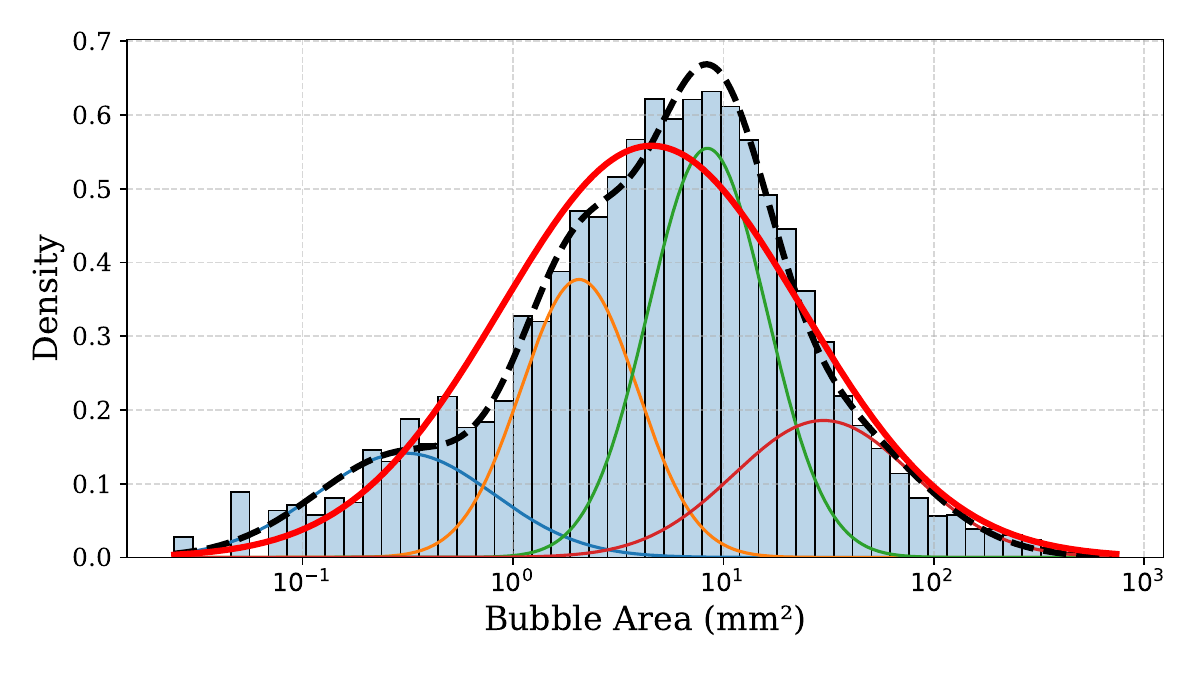}\vspace{-0.35cm}
    \caption{GMM fit to the bubble area distribution in log-scale. The single Gaussian (red) fails to represent the data, while the GMM (dashed black) captures its multi-modal structure. Colored components show bubble size clusters; values below the smallest cluster’s average were excluded as likely noise.}
    \label{fig:bub_dist_2}
\end{figure}

As already discussed in the introduction, past research focuses on well-shaped regular bubbles with limited variation in size and morphology. In this subsection, we briefly report on the characteristics of the proposed dataset to support the claim of a substantial increase in complexity.
%
%
In Figure~\ref{fig:size_comparison}, we compare the estimated probability density function on a logarithmic scale of our dataset against the two recent studies in the literature for which we could find extensive size data \cite{hessenkemper2022bubble, xu2024bubsam}. This dataset includes structures that are not only smaller but also significantly larger than those found in earlier studies. For example, the large end of the spectrum includes air patches, which are particularly relevant for drag reduction in air lubrication systems. 
Given the wide size variability, we performed a statistical analysis of bubble area distribution. Raw values showed no clear structure, so we applied a logarithmic transformation, which revealed a near-Gaussian trend. We then fitted a Gaussian Mixture Model (GMM), selecting the optimal number of components using the Bayesian Information Criterion. As shown in Figure~\ref{fig:bub_dist_2}, the single-Gaussian fit (red) captured the data poorly, while the GMM (dashed black) provided a more accurate representation. Later in our proposed data augmentation strategy, and the discussion of the results, we interpreted the individual components (thin colored lines) as meaningful clusters. The smallest component was likely due to noise or spurious dots, so we excluded values below its mean. In the rest of the paper, we refer to the remaining three clusters as small, medium, and large bubbles, with thresholds set at their intersection points. These include bubbles with areas ranging from below $1$ mm$^2$ up to $1000$ mm$^2$.


\subsection{Data Augmentation Strategy}

During preliminary evaluations of SAM 2.1 on the proposed segmentation, we quickly realized that large air patches were frequently observed and well detected. At the same time, medium and small bubbles were less accurately identified (quantitative assessment is available in the Results section). Thus, we decided to focus our data augmentation efforts on improving the model's performance for these smaller structures. During pre-processing, we auto-oriented the images and cropped them to retain only the region along the flow direction, corresponding to the upper half, where small and medium-sized bubbles are more concentrated. The images were resized to $640\times640$ pixels to ensure consistency.
Offline augmentations were performed at the mask level, directly modifying the segmentation masks to simulate realistic variations in the data. These included adding random noise affecting up to 0.1\% of pixels and applying shear transformations of up to $\pm10$ degrees horizontally and vertically around the bubble boundaries. 

In addition to these mask-level augmentations, standard image-level online augmentations were also used during training and runtime. These included random flipping and color adjustments, as specified in the SAM 2.1 training configuration file. This hybrid augmentation strategy increased the diversity of training data and enhanced the generalization capability.

\begin{figure*}[t]
    \centering
    \includegraphics[width=0.9\textwidth]{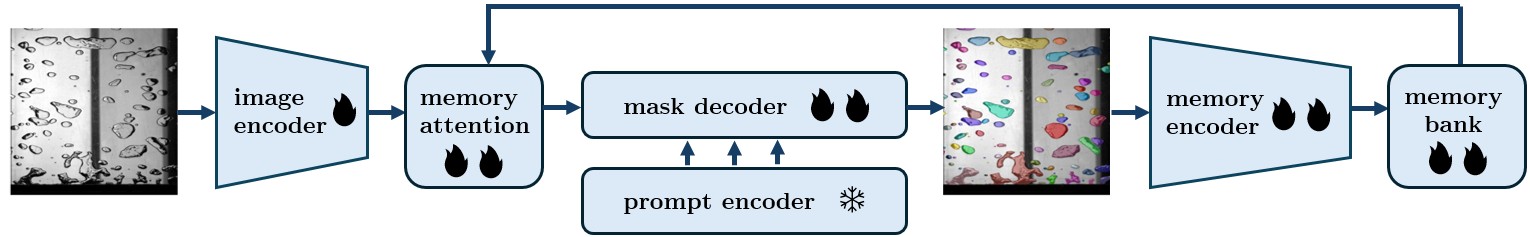}
    \caption{Architecture of the SAM 2.1 model used in our experiments. Modules marked with a snowflake were kept frozen, while those marked with a flame were fine-tuned. A single flame indicates partial tuning (e.g., only the final block of the image encoder), while two flames denote fully trainable modules such as the memory encoder and mask decoder.}
    \label{fig:arch}
\end{figure*}

\subsection{Metrics}

We define Precision and Recall as  $\mathcal{P} = {\mathrm{TP}}/{(\mathrm{TP} + \mathrm{FP})} $, $
\mathcal{R} = {\mathrm{TP}}/{(\mathrm{TP} + \mathrm{FN})} $, where TP and FP are true and false positives, and TN and FN are true and false negatives. Precision measures how many of the model's positive predictions are correct, while Recall indicates how many of the actual positive cases the model correctly identifies. F1 score combines precision and recall into a single harmonic mean, enabling balanced evaluation.
$\mathrm{F}1 = {2 \mathcal{P}\, \mathcal{R}}/{(\mathcal{P} + \mathcal{R})}
$. This metric is commonly used to balance the trade-off between precision and recall.

To evaluate segmentation quality, we report Intersection over Union (IoU) and Dice similarity, which quantify mask accuracy and boundary alignment. Dice, being more sensitive to overlap, is particularly informative for irregular or small structures. More precisely, IoU is defined as ${(A \cap B)}/{(A \cup B)}$, while Dice is ${2 (A \cap B)}/{(A + B)}$, where  A is the predicted mask and B is the ground truth mask.

\subsection{Fine-Tuning Strategy}

The fine-tuning process was carried out using the SAM 2.1 training framework, which comes with built-in tools for model development and adjustment. By offering ready-to-use settings for training steps, model configuration, and logging, it helps streamline the overall process. The framework also manages technical aspects, such as GPU usage, saving training checkpoints, and utilizing mixed-precision to accelerate computation. Thanks to its support for multi-GPU training with PyTorch’s DistributedDataParallel (DDP), it runs efficiently on both single systems and larger computing setups. More details on the training setup and code can be found in the official SAM 2.1 documentation \cite{ravi2024sam}.

The training configuration was defined in a separate YAML file, allowing precise control over which model components are trainable and how training parameters are set. In this setup, the image encoder was partially fine-tuned using a lower learning rate of $3 \mathrm{e}{-6}$ to retain pretrained features; specifically, only the trunk layers were updated, while the neck layers and embedding layers remained frozen to preserve their pretrained weights. In contrast, the mask decoder, memory attention, and memory encoder modules were fully trainable and used a higher learning rate of $5 \mathrm{e}{-6}$ to enable faster adaptation to the object segmentation task. This two-level learning rate setup keeps the general features steady while letting the task-specific parts learn faster. We also used a cosine annealing schedule to gradually lower the learning rates, which helped training be more stable and the model generalize better.

A multi-part loss combining cross-entropy, Dice, IoU, and classification losses was used, with extra weight on spatial overlap terms to better handle irregular object boundaries. Cross-entropy, being differentiable, allows effective training by reducing prediction errors, which indirectly improves precision and recall since these metrics are not directly optimized during training. The model was trained for 150 epochs using the AdamW optimizer, chosen for its better regularization and more effective handling of weight decay compared to standard Adam. The batch size was set to the maximum that fit in GPU memory—3 per GPU—and automatic mixed precision was enabled to reduce memory usage and speed up training. Regular augmentations like affine transforms, flipping, and color jitter improved generalization.

Optimizer settings, training duration, and logging intervals were all managed through the same config file to ensure reproducibility. The overall SAM 2.1 architecture is shown in Figure \ref{fig:arch}. More details and the training code are available on the project’s GitHub repository.

\section{Results}

To establish a reference point, the baseline performances of both the original SAM and SAM 2.1 models were first examined. SAM achieved an overall F1 score of 0.705, while the non-fine-tuned SAM 2.1 base model reached a slightly higher score of 0.720. Although SAM’s performance may initially seem competitive, a closer examination reveals certain limitations, particularly in the detection of large bubbles. While the model achieved a very high recall of $0.966$ for large bubbles, its precision was only $0.479$. This indicates that SAM often generated more than one mask for a single large bubble, leading to over-segmentation. Such errors can be critical in applications where quantifying the accurate number and size of bubbles is essential.
In contrast, the SAM 2.1 base model provided a more balanced performance for large bubbles, with a precision of 0.934 and a recall of 0.903. This improved trade-off between precision and recall highlights SAM 2.1 as the preferred candidate for the fine-tuning process in this study. 

\begin{figure}[t]
    \centering
    \includegraphics[width=0.45\textwidth]{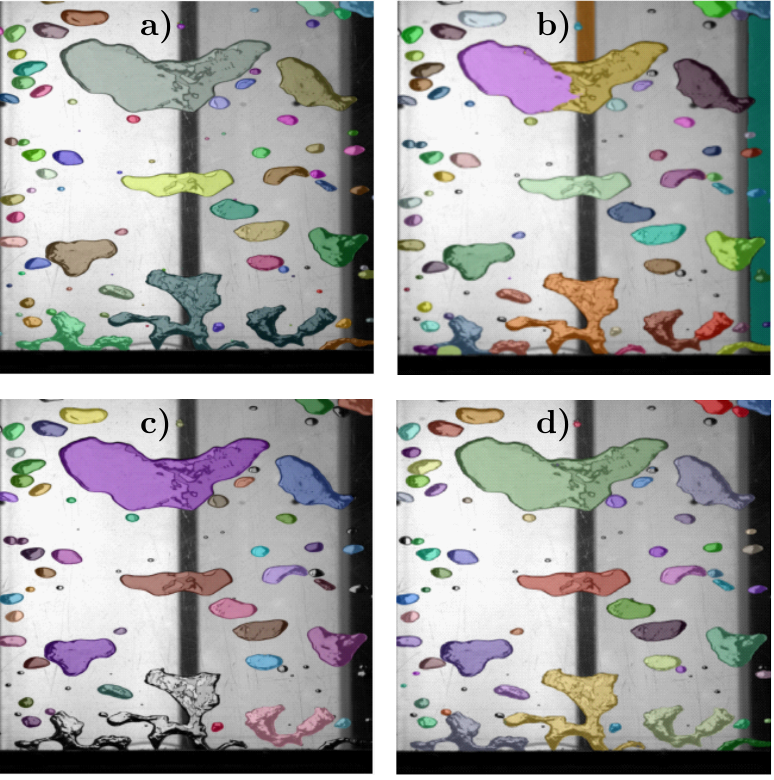}\vspace{-0.25cm}
    \caption{ Visual comparison of segmentation results. The air injector is located at the bottom of the images (shaded in black), and the flow is from bottom to top. (a) Manually annotated ground truth masks used as reference. (b) Output of the base Segment Anything Model (SAM). (c) Output of the base (i.e., non-fine-tuned) SAM 2.1 model. (d) Segmentation results from the fine-tuned SAM 2.1 model developed in this study, trained on 240 images using the Augmentation 2 strategy.}
    \label{fig:result2}
\end{figure}

This quantitative observation is supported by qualitative visual inspection of the masks produced by both models, superimposed on the raw images and compared against the ground truth annotations. As shown in Figure~\ref{fig:result2}, which reports the segmentation results for a representative frame, the baseline SAM model (Figure~\ref{fig:result2}b) successfully detects most of the bubbles and air patch regions. However, it also generates several spurious masks over background areas unrelated to the air phase—see, for instance, the shaded region on the left side of Figure~\ref{fig:result2}b.
Additionally, the model often splits individual objects into multiple adjacent masks (see the top of Figure~\ref{fig:result2}b) or creates overlapping ones for the same object (not easily visually distinguishable in the figure). These observations indicate that, while SAM exhibits high sensitivity, it lacks precision in delineating object boundaries and frequently produces masks that are irrelevant or redundant. In contrast, the SAM 2.1 base model (Figure~\ref{fig:result2}c) detects many air pockets while avoiding the generation of multiple masks for the same object, resulting in more coherent and consistent segmentations. Some irregularly shaped patches with ambiguous boundaries are missed, likely due to the model’s conservative prediction strategy, which only assigns masks when confident that a meaningful object is present. In this context, such caution is preferable to the over-segmentation seen in the earlier SAM version. Importantly, this behavior suggests that with appropriate fine-tuning on data containing sufficient air-phase exemplars, SAM 2.1 could achieve robust and reliable segmentation performance.

Based on these quantitative and qualitative findings, we selected the SAM 2.1 model as the foundation for fine-tuning in this study. Figure~\ref{fig:result2}d presents qualitative results from the fine-tuned model, trained on a dataset of 240 images (comprising 100 manually annotated and 140 augmented samples), alongside the baseline comparisons. The segmentation performance shows clear improvement: the model accurately captures the boundaries of air pockets—particularly medium and large ones—while only occasionally missing smaller bubbles. Detailed quantitative results and a description of the dataset construction are provided in the following section.

\begin{figure*}[htbp]
\centering
\makebox[0pt]{
\subfigure{\includegraphics[width=0.69\textwidth]{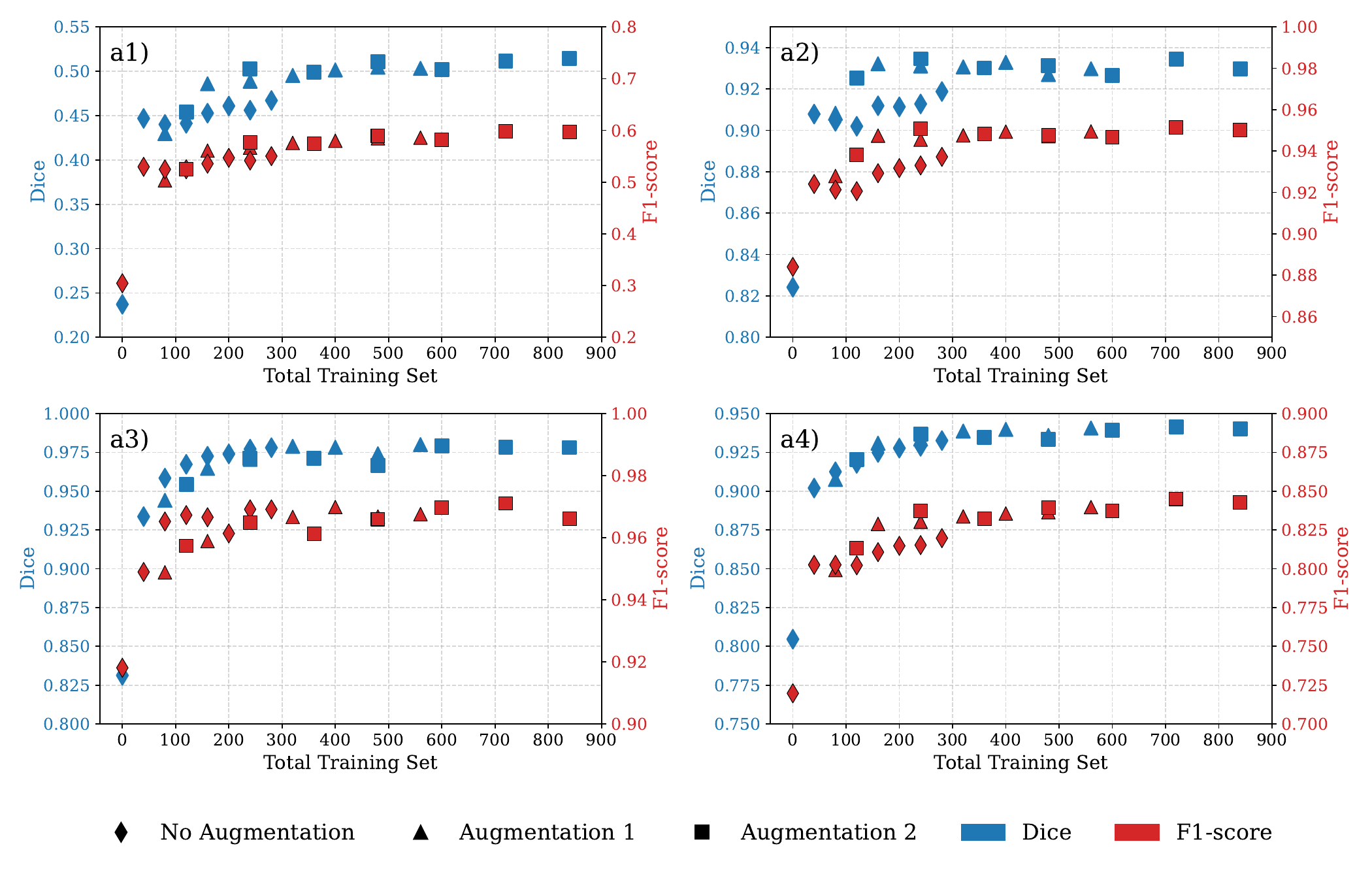}\label{fig:original}}
\hspace{0.01\textwidth}
\subfigure{\raisebox{9mm}{ 
\includegraphics[width=0.29\textwidth]{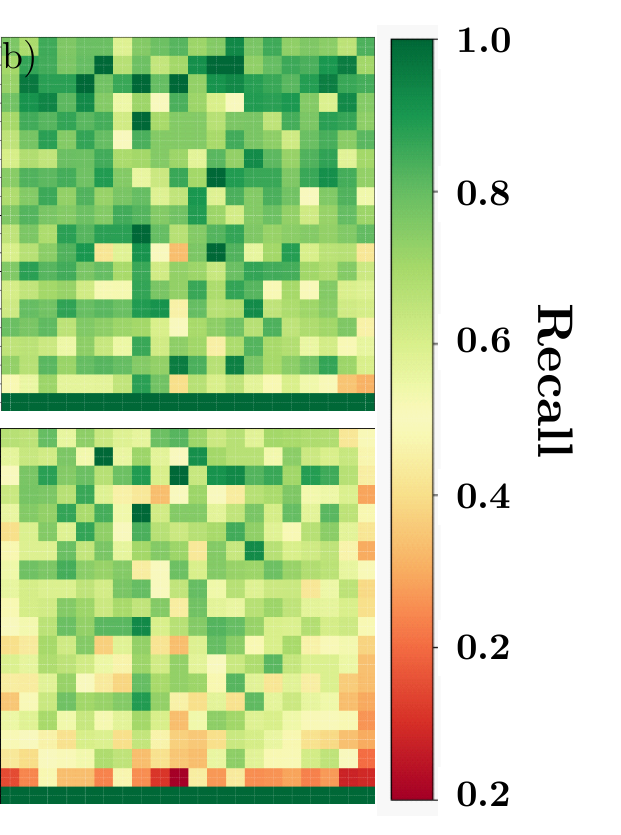}}\label{fig:segmentation}}
}\vspace{-0.25cm}
\caption{Segmentation performance of fine-tuned SAM2.1. (a) F1 and Dice scores for models trained on datasets of different sizes and with various augmentation strategies, shown for small (a1), medium (a2), and big (a3) bubbles (see section 2.2), and for all identified bubbles (a4). Evaluations are performed across three augmentation strategies: No Augmentation, Augmentation 1 (50$\%$ real and 50$\%$ augmented data), and Augmentation 2 (33$\%$ real and 66$\%$ augmented data).  (b) Spatial variation of model performance (based on recall values). Top: Fine-tuned SAM 2.1 using 240 training images with Augmentation 2 strategy. Bottom: Baseline SAM 2.1.}\label{fig:result}
\end{figure*}

To systematically evaluate the performance of the fine-tuned SAM 2.1 model and quantify the impact of data augmentation, we report F1 and Dice scores on a fixed validation set across three training sets of increasing size, each corresponding to a different augmentation strategy (Figure~\ref{fig:result}, left). The first set includes only manually labeled real images (No Augmentation), the second combines 50\% real and 50\% offline augmented images (Augmentation Rank 1), and the third uses one-third real and two-thirds augmented images (Augmentation Rank 2). This setup enables an empirical assessment of the trade-off between annotation effort (required for real images) and computational overhead (minimal for augmentations), and allows us to identify an optimal balance. For reference, we also compare these results with those from the baseline SAM 2.1 model (i.e., without fine-tuning, corresponding to a training set size of zero in Figure~\ref{fig:result}). Since our application involves segmenting images with highly variable bubble sizes, we report metrics both globally—across all detected bubbles (Figure~\ref{fig:result} a4)—and disaggregated by size category: small (Figure~\ref{fig:result} a1), medium (Figure~\ref{fig:result} a2), and large (Figure~\ref{fig:result} a3).

When looking at the overall metrics, for the smallest training set size tested (50 non-augmented images), the fine-tuned model already exhibits a 12$\%$ and 11$\%$ increase from the baseline case for Dice and F1 scores, respectively (Figure~\ref{fig:result} a4). As the training set size increases, performance keeps increasing monotonically (for all three augmentation strategies and both metrics) and then plateaus, beyond a set size of 240 images. A training dataset larger than that, irrespective of augmentation strategy, does not provide a meaningful performance improvement; thus for the rest of this section we will focus on results for training datasets of this size. This choice was also reflected in the visual result comparison above, where the output from the SAM 2.1 model, fine-tuned using 240 images with an augmentation of Rank 2 was included (Figure~\ref{fig:result2}d).

When comparing augmentation strategies, peak performance is very similar across all three, with F1 scores ranging from $0.815$ to $0.837$ and Dice scores reaching $0.929$ to $0.937$, steadily increasing from no augmentation to Augmentation Ranks 1 and 2. These results confirm that the fine-tuned model performs remarkably well under all tested configurations—achieving high-quality segmentation even when trained with limited manually labeled data. Two main conclusions follow from these observations. First, a training set composed of only 80 manually annotated images, complemented by augmented data, achieves performance on par with—or even slightly better than—that of a model trained on 240 real images. This demonstrates that data augmentation can significantly reduce annotation burden without compromising accuracy. Second, the marginal differences between Rank 1 and Rank 2 augmentation strategies suggest that the benefit of additional offline augmentation saturates, with limited gains beyond a certain point.

When the results are examined across different bubble size categories, performance is seen to vary significantly. Medium and large bubbles are segmented with high accuracy, exhibiting F1 scores between $0.95-0.965$ and Dice between $0.935-0.971$ (higher scores for the large bubbles). The performance metrics associated with large bubble sizes also differ slightly from the global ones discussed above, in terms of their variation with increasing training set size: performance does not monotonically increase but rather oscillates, especially when augmented images are used. This difference can be attributed to the fact that augmented data were created using only the top part of the original images, where mostly small and medium sized bubbles are present, thus somewhat limiting the benefits of augmentation for large bubbles, particularly for the F1 scores. Regardless, overall accuracy for these bubbles is the highest and is also seen to plateau around a similar training dataset size. In contrast, for small bubbles, although fine-tuning of the base model leads to almost doubling of both F1 and Dice scores, these still remain significantly lower, with values around 0.577 and 0.503, respectively. These results are chiefly due to our training choices, where lower weights were used for loss functions associated with small bubbles, given their relatively lower importance in the application of air lubrication. 

Finally, seeing that both illumination and bubble size distribution across the original images was non-uniform (Figure~\ref{fig:result2}), we also assess here whether there is any resulting spatial variation in model performance across the image plane. When looking at the recall values of the baseline SAM 2.1 model  (Figure~\ref{fig:result}b, bottom), we can see that there is indeed a spatial inhomogeneity present, with lower recall values close to the air injection, where larger, irregularly shaped air patches are present, and also on the left side, where illumination is insufficient. In contrast, the fine-tuned model (augmentation of Rank 2, for 240 training images) shows higher recall values overall, as expected, and no spatial dependence in performance (Figure~\ref{fig:result}b, top) with consistent segmentation across all regions, highlighting another gain from the fine-tuning process.


    

\section{Conclusion} 

This study demonstrated that high-quality bubble segmentation, across a wide range of sizes and shapes, can be achieved with minimal annotated data. Fine-tuning SAM 2.1 on as few as 100 labeled images (including the training and validation sets) led to substantial gains, particularly for medium and large bubbles, with F1 and Dice scores approaching 0.95. While small bubbles remain more challenging, this trade-off reflects an intentional bias in training priorities rather than a fundamental limitation. Compared to existing approaches that require extensive datasets yet struggle with generalization, our method offers a data-efficient alternative that remains robust in complex, real-world flow conditions. To support broader adoption, we release both our labeled dataset and fine-tuning pipeline—aiming to make accurate, low-effort segmentation accessible to the multiphase flow community.
Future work will focus on applying Fine-tuned SAM 2.1 as a tool in large-scale, multi-phase flow studies, as well as investigating bubble tracking.

\bibliographystyle{elsarticle-num} 
\bibliography{biblio}



\end{document}